# Standardization of the formal representation of lexical information for NLP

Laurent Romary, INRIA-Gemo & HUB-IDSL

Unter den Linden 6, 10099 Berlin

laurent.romary@inria.fr

## 1. Complexity of lexical structures and related domains

Lexical databases play a central role in all natural language processing applications (Briscoe, 1991), ranging from simple spellcheckers to more complex machine translation systems. In most cases, they constitute the sole parameter information for the corresponding software, and apart from some vary basic methods such as stemming (Lovins, 1968; Frakes, 1992) relying on pure string processing principles and low linguistic requirements, hardly any language technology application can avoid relying on a minimal lexical resource.

Even some basic tasks such as word segmentation for languages such as Japanese, Korean or Chinese (Halpern, 2008), in particular in the perspective of accurate named entity recognition, can hardly be carried out without large lexical resources. The same problem actually occurs for the proper identification of multi word units in "easier" languages as demonstrated in (Schone and Jurafsky, 2001). Such observations have lead ISO for instance to consider lexical representations as the main tenet of word segmentation processes (ISO/DIS 24614-1).

As a result, the cost of development and maintenance of a language technology application highly correlates with the complexity and size of the corresponding lexical database. It is thus essential to be able to standardise the structures and formats of lexical data, taking into consideration that the actual complexity and coverage may deeply fluctuate from one application to another.

As a matter of fact, lexical databases can cover many different levels, from simple morphosyntactic descriptions, like in the Multext framework (Ide and Véronis, 1994; Erjavec, 2004), up to multilevel lexical descriptions for machine translation (Lieske et alii, 2001). The degree of generalisation and factorisation in such lexica also impact on their reusability from one application to another and any standardisation effort related to lexical information should be able to cope with multiple types of combinations of linguistic description levels.

Finally, it would be difficult to speak about NLP lexica without eliciting the possible relation that these may bear with more human oriented resources. In this respect, we should acknowledge that machine readable dictionaries as well as terminological databases, even if conceived to fulfil other types of requirements, should not be seen as completely separated resources which would deserve unconnected standardisation activities.

Machine-readable dictionaries can indeed be related in two main aspects to NLP lexica. First, they are both more and more conceived on the basis of large corpus exploration methods to identify morphological, syntactic or semantic patterns. As a consequence, they naturally share similar information components (examples, statistical information, general form and sense organisation) that result from their common origin. Second, the prior existence of machine readable dictionaries in the digital world has made them a good basis for compiling NLP lexica by exploring their content (Copestake et alii, 1995). Despite the encountered difficulty to generate formal representations out of more prose based descriptions, such methods have proved useful to ensure that the target lexica bears a large linguistic coverage. As a consequence, descriptors and sometimes organisation principles for both types of lexical data, should be similar enough to ensure that such extraction or mapping operations can rely on the same principles and further that enough interoperability exists between them so that they can be transparently used when their actual ontological difference is not relevant.

We should also consider here how NLP lexica should relate to terminological databases. There again, we can identify usage scenarios where both types should be closely related. This is typically the case when the underlying NLP application has to deal with either multi-word units in texts, or with combinations of languages (translation or multilingual information retrieval). The interest here relies on the fact that huge terminological databases are continuously maintained by human experts or translators worldwide which make them, in the corresponding scientific or technical domains, the best reference for the identification or translation of the corresponding occurrences in texts.

## 2. Overview of foundational works

Early attempts at standardising lexical structures in the nineties have either focused on precisely defining the features needed for the basic description of lexical entries, as has been the case for the Multext lexical family, or on the definition of a generic structure for lexical databases, as in the TEI guidelines. We will come back below on the TEI representation when exemplifying LMF and we will first outline here why Multext has been an important step forward in the domain of standardisation.

The full specification (see Bel et alii, 1995) of the Multext descriptive framework for both morphosyntactic lexica and, by extension, for the morphosyntactic annotation of texts, is entirely based on flat feature structures, which in turn are encoded as elementary tags. For instance, the German form "Hundes" can, on the one hand, be abstractly characterised by the following set of features:

{cat=noun, type=common, gender=masculine, number=singular, case=genitive}

and, on the other hand, be encoded as a simple line representation in a text file as follows:

Hundes Hund Ncmsg

where the form is accompanied with an indication of the associated lemma together with a concise tag. The tag itself maps one-to-one onto the underlying feature structure so that features can univocally refer to the actual definition provided in the Multext specification. For instance, the specification describes the possible values for case in German as comprised of {nominative, genitive, dative, and accusative}, encoded in turn as {n,g,d,a} respectively in the tagset. As a matter of fact, the Multext framework implements the distinction between *conceptual domain* and *value domain* as defined in ISO 11179 and thus has paved the way for the current work on data categories (see below). Besides, the recent publication on a standardised XML format for feature structures (ISO 24610-1:2006) makes it actually possible to reuse Multext specifications and lexica in a modern technological framework (Erjavec, 2004).

The wide success of the Multext guidelines, in collaboration with the Eagles project, resulted from both their simplicity and comprehensiveness. They allowed the quick dissemination of reference resources in the initial seven languages of the project (English, French, Spanish, Italian, German, Dutch, Swedish), complemented by seven additional languages (Bulgarian, Czech, Estonian, Hungarian, Romanian, and Slovene, followed afterwards by Serbian) in the context of the Multext-east project (see http://nl.ijs.si/ME/). The intimate liaison with part of speech tagging paved the way to see such guidelines as a strong basis for interoperability and the feature-structure based formalism favoured the development of additional developments whereby new languages where being described under the same modelling framework.

From a more theoretical point of view, feature structures appear as a very appropriate framework for modelling the organisation of lexical structure and content. Without going back to the first shifts towards lexicalisation in the post-chomskian era, the expressive and inferential power of feature structures have been used in the context of highly lexicalized formalisms such as HPSG (Pollard and Sag, 1994) as well as general purpose lexical databases such as Acquilex (Briscoe, 1991). This relation has been made explicit up to the point of being seen as characterising the actual informational coverage of a lexical structure (Ide et alii 1993) as well as a way to elicit inheritance mechanisms existing within a lexical entry (Ide et alii, 2000). It may be noted that these concepts have been seminal in the design of the TEI Print dictionary chapter and LMF respectively.

A series of further developments took place in the late 90s and early years 2000 to a) extend the experience gained from the Multext project in standardising formats and associated semantics, b) integrate theoretical contributions in lexical modelling and c) bring in input from additional representation levels that gained maturity from a processing point of view.

In Europe a series of follow-up projects to Multext and Eagles, but also to pioneering SGML based activities such as Genelex (Antoni-Lay et alii, 1994), worked towards a comprehensive lexical model incorporating multiple linguistic levels. The Simple project (Bel et alii, 2000), followed by the cross-Atlantic Isle project (Calzolari et alii), designed a multi-layered architecture for lexical structures, which is sketched out in figure 1. What characterises this architecture is basically that multiple representation layers are both seen as autonomous modules, thus facilitating factorisation, and possibly linked together explicitly, thus ensuring a coherent framework for the development and maintenance of lexica. Both Simple and Isle produced multilingual lexical resources, which, despite their low public availability, provided useful examples for the further development of LMF.

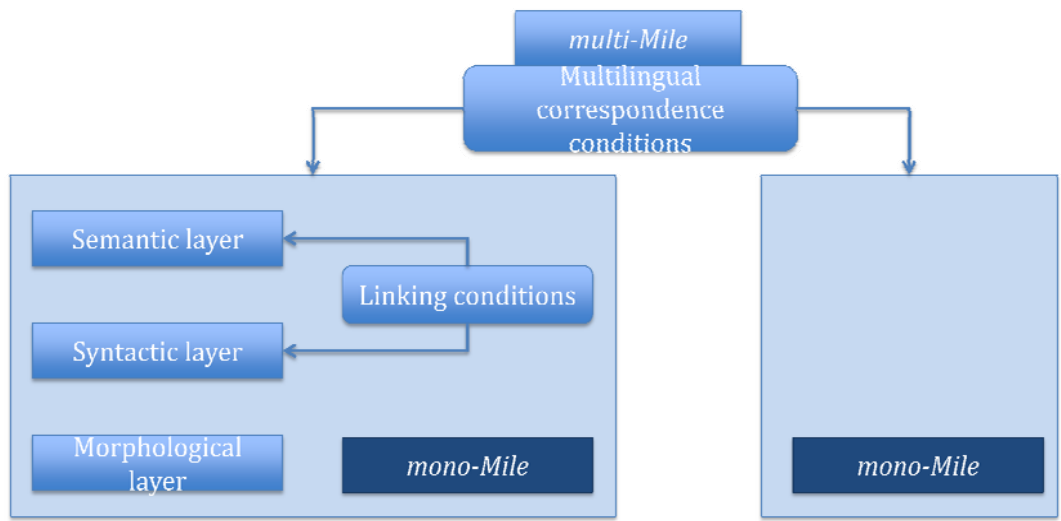

Figure 1: the general Mile lexical architecture.

In parallel to these initiatives aiming at providing large coverage lexica, we should mention here two specific endeavours that developed lexical resources for the general, somehow language independent, representation of linguistic semantic content, namely Framenet (Baker and alii, 2003) and Wordnet (Fellbaum, 1998). Although not directly aimed at being used in NLP applications, they had a strong influence on many projects dealing with lexical representation, semantic annotation or information retrieval from texts. The two initiatives are also somehow complementary to one another. Framenet offers, in the continuation of the works by Fillmore in knowledge representation, a surface annotation model for predicates and their participating components, which is factorised in a database of corresponding *frames*. The following example (taken from Ruppenhofer et alii, 2006) illustrates in particular how surface participants to the "frying" actions are characterised in close relation to the actual predicate.

[Cook Matilde] **fried** [Food the catfich][Heating_instrument in a heavy iron skillet]

Going in a quite complementary direction, Wordnet focuses on the broad semantic organisation of the lexicon by means of a general notion of *synset*, which groups together words having a close semantics. Synsets are in turn linked together by means of lexical or conceptual relations (meronymy, synonymy, antonymy, etc.).

Even if both initiatives cannot be seen as standards from the point of you of their data models they have *de facto* imposed some principles for the representation of lexical semantic information and again influenced deeply the design of LMF. The issue here is again clearly that of dealing with the variety and thus complexity of lexical structures and before we go any further in presenting LMF, let us go through the main trade-offs that have been at the basis of the design of the standard.

# 3. Intermezzo – which standardisation strategy for lexical resources

There are several issues that a standardisation organisation has to consider right from the onset before starting such an initiative as LMF. These have mainly to do with genericity and interoperability principles.

The first question at hand is whether one should standardise a specific (XML) format or on the contrary only state constraints as to what structures or models a lexical entry should follow in order to be compliant to the said standard. If the choice is to provide an actual format, it potentially facilitates immediate interoperability across applications, but bear the risk of not being flexible enough if some phenomena occur, that have not been anticipated in the standard. Conversely, focussing on the definition of abstract models may prevent implementers to find answers to their immediate needs in the standard proper, but offers the possibility to design an effective specification platform covering a wide range of potential applications.

The second question that one should ask himself when designing a new standard is how much it does fit the picture of existing standards and initiatives. This has to do with potentially already "existing" implementation of the standard (like the TEI), but also with other activities bearing a strong interface with the lexical domain.

A core example in the domain of Natural Language Processing is the morphosyntactic annotation of a text. As elicited in ISO standard 24611, morphosyntactic annotation consist in two stages, namely a tokenisation level and a word form identification level. Whereas the former is dedicated to segmenting elementary units on a language stream (written or spoken), the latter connects one or several tokens with a given entry within a lexicon. Because of such a link, it is essential that the descriptive principles used in the morphosyntactic annotations (e.g. grammatical descriptors) are coherent with those used in the lexical structures.

Another issue to be taken into account is the existence of widely disseminated XML vocabularies for representing lexical structures, in particular for big NLP lexica or dictionary projects. Like mentioned in the introduction separating one from the others would miss the prospect to pool together lexical sources or at least to develop similar tools (e.g. query engines) to manipulate them in a coherent way. A well-defined standard should thus find the best trade-off between being completely agnostic as to a specific syntax but all the same be able to cover a wide range of applications.

The designers of LMF have had these various aspects in mind and we shall try, in the remaining parts of this article to illustrate how these factors have been taken into account.

# 4. LMF

The LMF project was initiated in late 2003, following a request by the American delegation in ISO committee TC 37/SC 4 to have a flexible specification platform for lexical data, which would cope for multilingual and multilevel lexica. This idea was to work out a standard similar to the one developed for computerized terminologies (ISO 16642), i.e., providing a meta-model that can be further instantiated according to users' needs. To cope for this, a new working group, with Nicoletta Calzolari as convenor, was set up and a new project initiated with Gil Francopoulo and Monte George as editors: ISO 24613, Lexical Markup Framework, which was published in 2008.

LMF actually implements a fully-fledged *semasiological model*, i.e. a model that associates a word form to the variety of lexical descriptions that are considered as relevant for it. Extending from the usual situation of a human readable dictionary, where the descriptive target is the various meanings of a word, LMF also copes for other variations, such as syntactic constructs, whether or not they correlate with changes in meaning.

Under the general notion of word form, LMF puts together any information that document, classifies or structure the written or spoken representation of a lexical unit. For instance, the various transcriptions or transliterations (comprising phonetics) together with hyphenisation or syllabification information are recorded under the word form component, as well as multiple variants, possibly down to the description of inflected forms. In the case of multi-word units, the internal structure of the lexical item is also part of such a representation.

The sense component is organised as a fully iterative and recursive structure, which, according to the scope of the actual lexical database to be implemented may be further characterised by any number of constraints, e.g. register, usage, grammatical or syntactic variants, collocations or translations.

Further descriptive levels are made available in LMF in the form of additional, yet optional, packages that a lexicon developer may use to describe a model that matches his needs. These packages, as they stand in the published ISO standard, can be classified into three main categories:

- packages providing additional descriptive tools at morphological level, namely for representing morphosyntactic paradigm, multi word paradigms or MRD specific features;
- packages for NLP application, covering the syntactic, semantic and multilingual domains;
- a general package for expressing constraints on lexical entries.

All these packages may be used in combination with the single mandatory one, the *core package*. As depicted in figure 2, the core package provides the components for representing overall structure of a lexical database. Both at the macro-structural and meso-structural levels, the meta-model is fully hierarchical, with a *lexical resource* component grouping together a header (*Global information*) and one or several *Lexica*. Each lexicon, whose semantics depends on the editorial choices of the implementer (e.g. monolingual components in a multilingual dictionary, or specific gathering of full form information independently of sense descriptions), groups together a set of lexical entries. Finally, a lexical entry is characterised by at least one *form* component, possibly further refined by so-called *form representations* (e.g. for orthographical variants), together with any number of *senses*.

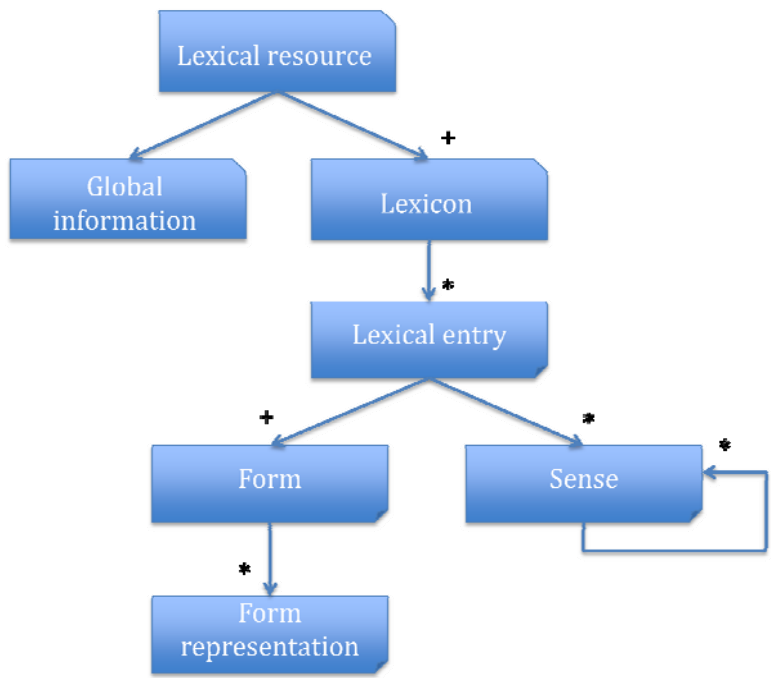

**Figure 2: the LMF core package**

On the basis of the components defined in the LMF standards, it is rather straightforward to design a concrete XML representation compliant with the core package. Since the core package represents a hierarchy, just as XML information structure does (cf. article 93), one can simply associate each component to an XML element and further complement the skeleton thus obtained by additional elements representing the data categories that one want to associate to each level. As an illustration, we show in figure 3 an entry taken from the Morphalou project (Romary et alii, 2004), which provides a full form lexicon for the French language. As can be seen the Lexical entry, Form and Form representation components are instanciated as <lexicalEntry>, <formSet> and <lemmatizedForm> or <inflectedForm> (to differentiate between lemma and inflexion). Sense is not present here, since it is not relevant for such simplified forms of lexical data.

```
<lexicalEntry xml:id="championne_1">
<feminineVariantOf target="#champion_1">champion</feminineVariantOf>
<formSet>
      <lemmatizedForm>
            <orthography>championne</orthography>
            <grammaticalCategory>commonNoun</grammaticalCategory>
            <grammaticalGender>feminine</grammaticalGender>
      </lemmatizedForm>
      <inflectedForm>
            <orthography>championne</orthography>
            <grammaticalNumber>singular</grammaticalNumber>
      </inflectedForm>
      <inflectedForm>
            <orthography>championnes</orthography>
            <grammaticalNumber>plural</grammaticalNumber>
      </inflectedForm>
</formSet>
</lexicalEntry>
```
**Figure 3: example of an LMF compliant entry from the Morphalou project**

# 5. Standardising descriptors

In the preceding sections we have given some hints as to the importance of, on the one hand, providing clear models that act as an interoperability framework across applications, but also, on the other hand, identifying reference formats to facilitate the development of generic tools as well as the quick dissemination of data. Still, no real standardisation objective is entirely fulfilled whether at model, nor at format level, if we miss mechanisms to guaranty the consistency of the semantics associated with the information conveyed by a lexical structure.

One solution to this problem is provided by the notion of *data category*, which stabilizes the semantics of elementary fields within linguistic structures (Ide & Romary 2004a). Data categories are equivalent, in the linguistic field, to the notion of *data element concept* in ISO 11179-3. They actually play a double role:

- They provide unique reference points for computer applications (schema definitions, software interfaces, query languages) to refer to, with the objective to ensure a one-to-one mapping across various data representations of the same underlying concept (e.g. /feminine/ being implemented as *f* in one format and *fém.* in another);
- They are documentary objects that provide definitional semantics for human users, so that they can differentiate, within a given field, which data category actually matches the concept that he or she has in mind (say /definite determiner/ vs. /indefinite determiner/).

ISO committee TC 37 has thus been working, in the context of the development of the ISO 12620 standard, on the definition of a specific application of data categories to language resources. In particular this has lead to the identification of an important trade-off:

- Providing generic entry points and corresponding unique identifiers for concepts which are widely shared across language technology application, independently, or much so, of the actual language to be processed;
- Designing data categories in such a way that fine grained information about a linguistic concept which is only relevant for one or a reduced number of languages (e.g. the possible values for /gender/) is actually recorded.

This is indeed the way ISO 12620 has been defined and further implemented in the reference ISOCat *data category registry* (see http://www.isocat.org). This endeavour actually takes up the expertise gathered-up in such previous projects as EAGLES to make it a fully operational infrastructure.

# 6. Standardising the syntax – convergence with TEI

The last stage in the direction of offering the language technology community with a comprehensive standardisation picture for representing lexical structures is to analyse how one achieving a better convergence between existing endeavours focusing either on the modelling or on the actual concrete representation of lexical data, so that these can be seen as two complementary answers to implementers' needs.

As mentioned in the introduction of this chapter, the TEI has been a major initiative in the last twenty years in providing standards for the humanities computing community and in particular in the domain of dictionary representation. Not only were the TEI guidelines for dictionary encoding intended to provide representational means for print dictionaries (also known as *machine readable dictionaries*), but also for basic lexical structures usable in NLP applications. It is thus worth considering if the TEI can be a reference syntax for a subset of LMF and beyond, if it can be taken as a basis for providing formats with a wider compliance to the various LMF extensions. We will not deal here with this second more prospective issue, even if we could mention that the customisation mechanisms of the TEI, with the ODD specification language, provide all technical means to actually go in this direction.

Let us focus here on how the TEI can be used for representing LMF compliant full form lexica. To this end, we can first identify the core components (or *crystals*) that the TEI offers for lexical description.

The general organisation of a TEI dictionary <entry> is actually based on a semasiological structure grouping together, just like in LMF, two main groups of information:

- information about the morphological and grammatical characteristics of the entry, by means of the elements <form> and <gramGrp> respectively;
- a recursive element <sense> where all restrictional and semantic information are grouped together;

Still, the general TEI model definition allows any iteration, as well as recursive combination, of <form> and <gramGrp>, which makes the TEI model potentially depart from the more restrictive LMF core and morphological packages. To circumvent this, we need to identify the best standard match between the TEI objects and the LMF components by both simplifying and at times over-specifying the former accordingly.

Without fully eliciting here the mapping process, we can see that <form> clearly matches the Form component in LMF and may be further typed to account for the three instance components: Lemma, Word form and Stem. Its recursive nature can also be used to group together inflected forms when necessary. As an example, we can take up an excerpt from the preceding Morphalou entry we had for "champion" to make it TEI compliant as follows:

```
<entry>
        <form type="lemma">
                <orth>championne</orth>
                <gramGrp>
                        <pos>commonNoun</pos>
                        <gen>feminine</gen>
                <gramGrp>
        </form>
        <form type="inflected">
                <orth>championne</orth>
                <num>singular</num>
        </form>
        <form type="inflected">
        <form type="inflected">
                <orth>championnes</orth>
                <num>plural</num>
        </form>
        </inflectedForm>
</entry>
```
**Figure 3: an LMF compliant TEI entry**

Among the restrictions that one has to apply to the TEI model we can mention the limitation of recursivity for the form element or the constraining of the type attribute of form to a set of LMF compliant values.

## 7. Perspectives – even more convergence

The initial design of a lexical database has long-term impact on the capacity that one has to edit, maintain, exploit or disseminate the corresponding content. Therefore, the issue of the underlying lexical model, and thus data format, should be taken extremely seriously without considering it as something too technical to be endowed with the adequate effort and level of expertise and at the end of the day left to pure IT experts.

Even if it requires one to acquire, beyond the sole scientific knowledge in NLP or lexicography, additional competences as to the existing standardisation initiatives, we need to get the computational linguistic community to master a real culture of specification in the lexical domain, but also at large in the representation of linguistic resources. This is probably the only way to go towards a global interoperability between the various language resource repositories and tools that have been developed over the years and in turn, to make sure that the actual energy in computational linguistics is spent on designing new methods or achieving new scientific results, rather then maintaining ad-hoc technical infrastructures.

## 8. Normative references

ISO 16642:2003, Computer applications in terminology — Terminological markup framework

ISO 24610-1:2006, Language resource management — Feature structures — Part 1: Feature structure representation; also available as a module of the TEI guidelines (see http://www.tei-c.org/release/doc/tei-p5-doc/en/html/FS.html)

ISO/DIS 24611, Language resource management — Morpho-syntactic annotation framework

ISO 24613:2008, Language resource management — Lexical markup framework (LMF)

ISO/DIS 24614-1, Language resource management -- Word segmentation of written texts for monolingual and multilingual information processing -- Part 1: Basic concepts and general principles

ISO/IEC 11179-3, Information Technology -- Metadata registries (MDR) – Part 3: Registry metamodel and basic attributes

## 9. Select bibliography